\pdfoutput=1
\documentclass[11pt]{article}

\usepackage[final]{acl}

\usepackage{times}
\usepackage{latexsym}

\usepackage[T1]{fontenc}
\usepackage[utf8]{inputenc}

\usepackage{microtype}

\usepackage{inconsolata}

\usepackage{graphicx}
\usepackage{kotex}
\usepackage[utf8]{inputenc}
\usepackage[T1]{fontenc}
\usepackage{natbib}
\usepackage{array}
\usepackage{booktabs}
\usepackage{multirow} 

\usepackage{graphicx}
\usepackage{kotex}
\usepackage[utf8]{inputenc}
\usepackage[T1]{fontenc}
\usepackage{natbib}  % 이 줄 추가!

\usepackage{array}
\usepackage{booktabs}
\usepackage{multirow}

\title{KoGEC : 
Korean Grammatical Error Correction with
Pre-trained Translation Models
}

\author{Taeeun Kim\textsuperscript{1,2} \ Semin Jeong\textsuperscript{1} \ Youngsook Song\textsuperscript{1} \\
\texttt{taeeunk1208@sionic.ai} \ \texttt{hm@sionic.ai} \ \texttt{song@sionic.ai} \\
\textsuperscript{1}Sionic AI Inc., Seoul, Korea \\
\textsuperscript{2}Emory University, Atlanta, GA, USA}

\begin{document}

\maketitle
\begin{abstract}
This research introduces KoGEC, a Korean Grammatical Error Correction system using pre\--trained translation models. We fine-tuned NLLB (No Language Left Behind) models for Korean GEC, comparing their performance against large language models like GPT-4 and HCX-3. The study used two social media conversation datasets for training and testing. The NLLB models were fine-tuned using special language tokens to distinguish between original and corrected Korean sentences. Evaluation was done using BLEU scores and an "LLM as judge" method to classify error types.
Results showed that the fine-tuned NLLB (KoGEC) models outperformed GPT-4o and HCX-3 in Korean GEC tasks. KoGEC demonstrated a more balanced error correction profile across various error types, whereas the larger LLMs tended to focus less on punctuation errors.
We also developed a Chrome extension to make the KoGEC system accessible to users. Finally, we explored token vocabulary expansion to further improve the model but found it to decrease model performance.
This research contributes to the field of NLP by providing an efficient, specialized Korean GEC system and a new evaluation method. It also highlights the potential of compact, task-specific models to compete with larger, general-purpose language models in specialized NLP tasks.
\end{abstract}

keywords :
{Korean, Grammatical Error Correction, NLLB, LLM as a Judge}

\section{Introduction}

Korean, like many languages, lacks validated Grammatical Error Correction (GEC) models. This gap is particularly significant given the complexity of Korean grammar, which poses unique challenges due to its agglutinative structure, extensive particle system, intricate word spacing rules, and complex verb conjugations. These factors make it difficult even for native speakers to write grammatically correct Korean, highlighting the need for automated correction systems. 

This study aims to establish a language model that prioritize the preservation of the author's original intent while correcting grammatical errors and typographical mistakes, moving away from sentence paraphrasing. Our proposed model, NLLB\_ko\_gec, is based on the NLLB (No Language Left Behind), a multilingual model capable of translating between 200 languages \citep{nllbteam2022languageleftbehindscaling}.  Building upon the work of \citep{luhtaru-etal-2024-no}, who leveraged NLLB models for multilingual and low-resource Grammatical Error Correction (GEC), this study expands the language coverage beyond their initial focus on English, Czech, and German. We extend the application of multilingual machine translation (MT) models to Korean GEC, incorporating a language with a distinct writing system to further explore the versatility of NLLB in automated error correction across diverse linguistic contexts. We perform a comparative analysis of the automated Korean GEC performance of state-of-the-art models such as GPT-4o with OpenAI and HCX-3 with Naver Cloud against the NLLB\_ko\_gec model.
Furthermore, this research seeks to contribute to the advancement of the open-source community by publicly releasing the developed research findings under a CC\--BY\---NC license. 

Systematic error classification, such as the 28 error types proposed in the \citep{ng-etal-2014-conll}, is crucial for understanding the characteristics of individual languages and identifying commonalities between languages. In this study, we have systematized Korean specific error types through collaboration between linguists and computer scientists. By examining grammatical error correction and error types across the Korean language group, we hope to address the specific challenges of Korean grammar.

For performance evaluation, we applied the 'LLM\--as\--judge' method proposed by \citet{zheng2023judging} and followed the \citet{korean-spelling-criteria}.

With the rapid advancement of Large Language Models, there is growing recognition of the importance of training data quality for these models. Grammar and spelling verification are essential in the data quality inspection process, and this study proposes an automated quality checking mechanism utilizing the fine tuned NLLB\_ko\_gec model. 

The significance of grammatical error correction extends beyond data quality inspection; it is crucial for effective communication in academic, professional, and social contexts. Thus, we also propose a Chrome extension service that demonstrates NLLB\_ko\_gec’s impact on various social aspects of communication and accessibility.

\section{Related Work}

Grammatical Error Correction (GEC) has been an important task in the natural language process for a long time. With the emergence of ChatGPT, there have been studies aimed at verifying whether it can improve GEC performance on datasets such as the CoNLL-2014 Shared Task on Grammatical Error Correction (\citet{ng-etal-2014-conll}) and hybrid datasets for English, German, and Chinese. One such study is by \citet{wu2023chatgpt}. In their research, \citet{wu2023chatgpt}. compared the GEC performance of ChatGPT and Grammarly. For long sentences, the recall scores were 62.8 for ChatGPT and 45.3 for Grammarly, indicating that both systems failed to achieve satisfactory scores. Additionally, it was observed that ChatGPT tends to rephrase sentences, which deviates from the original intent of GEC that primarily focuses on minimizing edits as a key evaluation criterion.

While ChatGPT's rephrasing increases the overall fluency of the input sentences, it often results in semantic variants or changes in voice and style. Recognizing the distinction between grammatical error correction and general writing assistance, users who simply want to correct grammatical errors may not want a model to arbitrarily change their writing. Therefore, controllability should be considered a crucial requirement for using ChatGPT in GEC applications.
To address these limitations, researchers have explored alternative approaches. Previous works have suggested that Machine Translation (MT) models can be effective in grammatical error correction tasks by treating the conversion of erroneous sentences to correct sentences as a translation task. This methodology has led the field to adopt single-direction MT models for GEC, successfully implementing neural techniques for GEC system development.
In the context of the Korean language specifically, \citet{yoon-etal-2023-towards} and \citet{Maeng2023EffectivenessOC} developed Korean grammar error categorizations. However, these research studies do not solely focus on native Korean speakers; their primary emphasis is on Korean language learners. Consequently, among the error types categorized, one can observe categories for errors that native Korean speakers rarely make.

In the following examples mentioned in \citet{yoon-etal-2023-towards}, in 'An error on ending', the correction from '나무 (tree)' to '너무 (too)' was made, and in 'CONJ An error on conjugation', '잘라에 (to Zalra, place)' was corrected to '자르러 (to get my haircut, purpose)'. Such errors are unlikely to be regular or frequent mistakes made by native Korean speakers who use an agglutinative language as their mother tongue. Therefore, in this study, we created guidelines based on the Korean spelling evaluation criteria as per the Ministry of Culture, Sports, and Tourism Notice No. 2017/-12 (March 28, 2017).

\section{Data Collection}
Our primary objective in developing a Korean GEC system was to address grammar errors made by native Korean speakers, rather than those of Korean language learners. This focus was chosen because native speakers' errors are typically more straightforward and context-specific. In contrast, learners' mistakes often involve ambiguities in intended meaning, making them more susceptible to misinterpretation and inadvertent paraphrasing during the correction process. 

We utilize two native conversation datasets provided by government-supported institutions. The first dataset is the NIKL Spelling Correction Corpus 2021, provided by the National Institute of Korean Language in 2022\footnote{(Source) National Institute of Korean Language (2022). NIKL Spelling Correction Corpus 2021 (v.1.0). URL: kli.korean.go.kr.}. The second dataset is the Korean Error Correction Data 2023, provided by the National Information Society Agency\footnote{(Source) National Information Society Agency (2023). Korean Error Correction Data. URL: www.aihub.or.kr.}.

The first dataset was collected from social media conversations and was propagated with emojis and data in English. Thus, our pre-processing involved replacing emojis with empty strings and removing data that only held English. Additionally, for the second dataset, we removed voice recognition error correction data which contained corrections that altered the sentences' meanings entirely. A section of this dataset labeled as '오탈자 데이터(typo dataset)' held data with identical correct sentences corresponding to slightly different error sentences. This was discarded due to concerns of overfitting. After preprocessing and concatenation, our final training dataset consisted of approximately 520k rows in total.

\begin{table}[h]
\centering
\begin{tabular}{|c|c|c|}
\hline
\textbf{Corpus} & \textbf{Train} & \textbf{Test} \\
\hline
NIKL Spelling \newline Correction Corpus & 393k & 4k \\
\hline
Korean Error \newline Correction Data & 127k & 1k \\
\hline
\textbf{Total} & \textbf{520k} & \textbf{5k} \\
\hline
\end{tabular}
\caption{Corpus Statistics}
\label{tab:corpus-size}
\end{table}

Building on previous works, our study aimed to explore the potential of compact translation models in Korean GEC tasks. While it is intuitive that larger language models like LLaMA or Mixtral, with their vast parameter counts and extensive training data, would yield superior results, the objective was to minimize compromising performance quality while using smaller, task-specific models designed for low-resource environments. 
We selected the No Language Left Behind (NLLB) model for our primary experiments. NLLB, a compact yet specialized translation model capable of translating 200 different languages, aligned with our research objectives for reasons below:
\medskip
\begin{itemize}
    \item \textbf{Specialized Architecture}: As a translation model, NLLB demonstrates superior grammatical parsing and generation capabilities compared to general-purpose language models of similar size. This specialization is particularly advantageous for GEC tasks, which require nuanced understanding and manipulation of grammatical structures.
    \item \textbf{State-of-the-Art Performance}: Among translation models in its class, NLLB exhibits state-of-the-art performance. This characteristic makes it an ideal candidate for pushing the boundaries of GEC performance within the constraints of smaller model sizes.
    \item \textbf{Efficiency}: By choosing 600M and 3.3B parameter models over larger alternatives, we aim to demonstrate that efficient, task-specific models can compete with or outperform more resource-intensive general-purpose LLMs in specialized tasks like GEC.
        \end{itemize}

Our focus on compact models is driven by the imperative for computational accessibility and the democratization of AI technologies. By prioritizing efficiency and specialization, we aim to demonstrate that state-of-the-art performance in specific NLP tasks, such as grammatical error correction, can be achieved without the extensive computational resources required by large language models.

\section{Experiments}
\subsection{Dataset split}

The total number of rows in the dataset was 525,268, with 520,015 rows used for training and 5,253 for testing. The test dataset was further refined to remove data irrelevant to grammatical error correction (GEC) tasks, such as rows containing only strings of repeated Korean characters like "ㅋㅋㅋㅋㅋㅋ." These expressions are often used in Korean text to mimic laughter or express amusement, similar to "haha" in English. The dataset included two main columns: ‘original form’ and ‘corrected form.’ The ‘original form’ column contains Korean sentences with various grammatical errors, while the ‘corrected form’ column provides the grammatically correct versions of these sentences.

\subsection{Model Training}
One of the techniques the NLLB model utilizes to translate between numerous languages is through special language tokens. Instead of a <bos> token, the NLLB uses language tokens that specify the beginning of a specific language. For example, Korean is designated by the <kor\_Hang> token. For the models to recognize the correction process from the original to the corrected form of a Korean sentence as a type of translation, we added a special token, <cor\_Hang> to identify the correct sentence. Although this process is not necessary, we observed a much better susceptibility to the GEC task when we distinguished between the two types of data.
We fine-tuned the NLLB model with the Adafactor optimizer (\citet{shazeer2018adafactor}). We utilized a single NVIDIA A100 GPU, setting batch sizes of 64 and 16 for the 600M and 3.3B models, respectively, with an update frequency of one. A constant learning rate scheduler with warm-up was implemented, performing warm-up for the first 1,000 updates.

The maximum sequence length was set to 128 tokens to accommodate our dataset of sentence pairs. Training data consisted of original and corrected sentence pairs, with batches generated by randomly selecting two language pairs.

The entire fine-tuning process spanned approximately 13 hours: the 600M model took 3 hours, while the larger 3.3B model required 10 hours. During training, we monitored the average loss every 200 steps and saved model checkpoints every 2,000 steps. The best checkpoint was selected based on performance on a development set.

\section{Results}
\subsection{Evaluation and Comparison}
Our experiments yielded two models, NLLB-200-ko-gec-3.3B and NLLB-200-ko-gec-600M that were derived from fine tuning two of meta’s open source models, NLLB-200-3.3B and NLLB-200-Distilled-600M. We compared the two resulting models to large, general-purpose LLMs:  GPT-4o and HCX-3. Currently, these two models are evaluated to have one of the best model performances in Korean (\citet{yoo2024hyperclovaxtechnicalreport}). Specifically, HCX-3 is a result of an effort to create an LM tailored to Korean language and culture by Naver Cloud’s AI team.  It has been reported that a third of HCX-3’s pre-training data consists of Korean, with the rest being multilingual and code data. The technical report states that HCX-3 and GPT-4o show comparable performance in translations between Korean and English.

% \begin{document}
% \begin{table}[h]
%     \centering
%     \begin{tabular}{|L|L|c|c|c|}
%         \hline
%         & \multicolumn{2}{c|}{\textbf{NLLB-200}} & \textbf{GPT-4o} & \textbf{HCX-3} \\
%         \cline{2-3}
%         & \multicolumn{2}{c|}{\textbf{ko-gec}} & & \\
%         \hline
%         & \textbf{3.3B} & \textbf{600M} & &\\
%         \hline
%         \textbf{BLEU} & 85.73 & 58.15 & 75.03 & 71.24 \\
%         \hline
%     \end{tabular}
%     \caption{Comparison of BLEU Scores}
% \end{table}
% % \end{document}

\begin{table}[h]
    \centering
    \begin{tabular}{|l|l|c|c|c|}
        \hline
        & \multicolumn{2}{c|}{\textbf{NLLB-200}} & \textbf{GPT-4o} & \textbf{HCX-3} \\
        \cline{2-3}
        & \multicolumn{2}{c|}{\textbf{ko-gec}} & & \\
        \hline
        & \textbf{3.3B} & \textbf{600M} & &\\
        \hline
        \textbf{BLEU} & 85.73 & 58.15 & 75.03 & 71.24 \\
        \hline
    \end{tabular}
    \caption{Comparison of BLEU Scores}
\end{table}

We assessed each model via BLEU (Bilingual Evaluation Understudy) scores\footnote{Once the test dataset was used for inference, the output was normalized properly. We found that the test dataset represented 
single Korean characters, such as 'ㅎ' and 'ㅋ' that is used as a 
consonantal expression similar to 'LOL' in English, with Hangul Compatibility 
Jamo Unicode. In contrast, the model outputs were expressed with Hangul 
Jamo Unicode. The differences in Unicode interfered with producing an 
accurate analysis of the results. We found an increase in BLEU scores after 
the normalization process across all models, with an increase as high as 
3.12 in the NLLB-200-ko-gec-3.3B model. We conclude that when replicating experiments in Korean, it is essential to verify Unicode normalization.} 

Since the metric compares model outputs to human-translated reference text, we determined that it would be appropriate to judge GEC quality as well. In this paper, we utilize the BLEU scores for all general performance examinations in reference to the correct data. 
Both LLMs, GPT-4o, and HCX-3, were initially tested using GEC instructions and a comprehensive guideline detailing standard Korean grammar rules (see appendix B). Each section of the guideline was accompanied by examples. To assess the effectiveness of the guideline and evaluate the general understanding of Korean grammar by GPT and HCX, we compared these results with those obtained using a zero-shot, instruction-only prompting method.

The comparison revealed minimal differences between the two approaches, leading us to conclude that both language models had acquired a respectable level of knowledge about the Korean language and its grammar through their pre-training processes. While the few-shot guidelines did slightly enhance the models' GEC capabilities, we determined that this improvement didn't justify the increased token input required. Consequently, we opted to conduct our final model evaluations using the zero-shot prompting method.

As shown in Table 2, the NLLB-200-ko-gec-3.3B model achieved a BLEU score of 85.73, substantially higher than the scores of 75.03 and 71.24 for GPT-4o and HCX-3, respectively. The superior performance of our ko-gec models demonstrates their effectiveness and potential for practical applications in Korean language correction and editing tools.

\subsection{LLM as a Judge}

To further investigate the fine-tuned models and their capabilities, we designed an annotation metric that utilizes an LLM as a Judge. We had researchers visually inspect the results of the LLM as a Judge to further identify and validate limitations for future improvements. With the main focus of getting a comprehensive view of each GEC system's limitations for later improvements, we constructed a classification of Korean grammar error types (see appendix A). We then prompted the LLM to inspect each GEC model's inference data outputs to determine the types of grammar errors they failed to catch. The classification was based on the category of error types proposed by \citet{yoon-etal-2023-towards}, which distinguishes Korean’s unique linguistic characteristics in 14 different error types, labeling them with error codes and examples. These categories were reduced to 11 error types by researchers, as a few of them were identified as error types only applicable to Korean learners, not natives. We chose GPT-4o to judge the types of errors within output data based on the criteria and print its error codes. To minimize errors, we implemented the reference-guided grading method suggested in previous research, where the LLM judge is provided with a reference solution to compare the model’s answer with. This method provides a clear benchmark for judging, minimizing self-enhancement bias and bypassing the issue of GPT-4o’s limited grading capability (\citet{zheng2023judging}). The generated set of error codes was compiled to study the prevalence of each type. 
\begin{figure*}
    \centering
    \includegraphics[width=330pt]{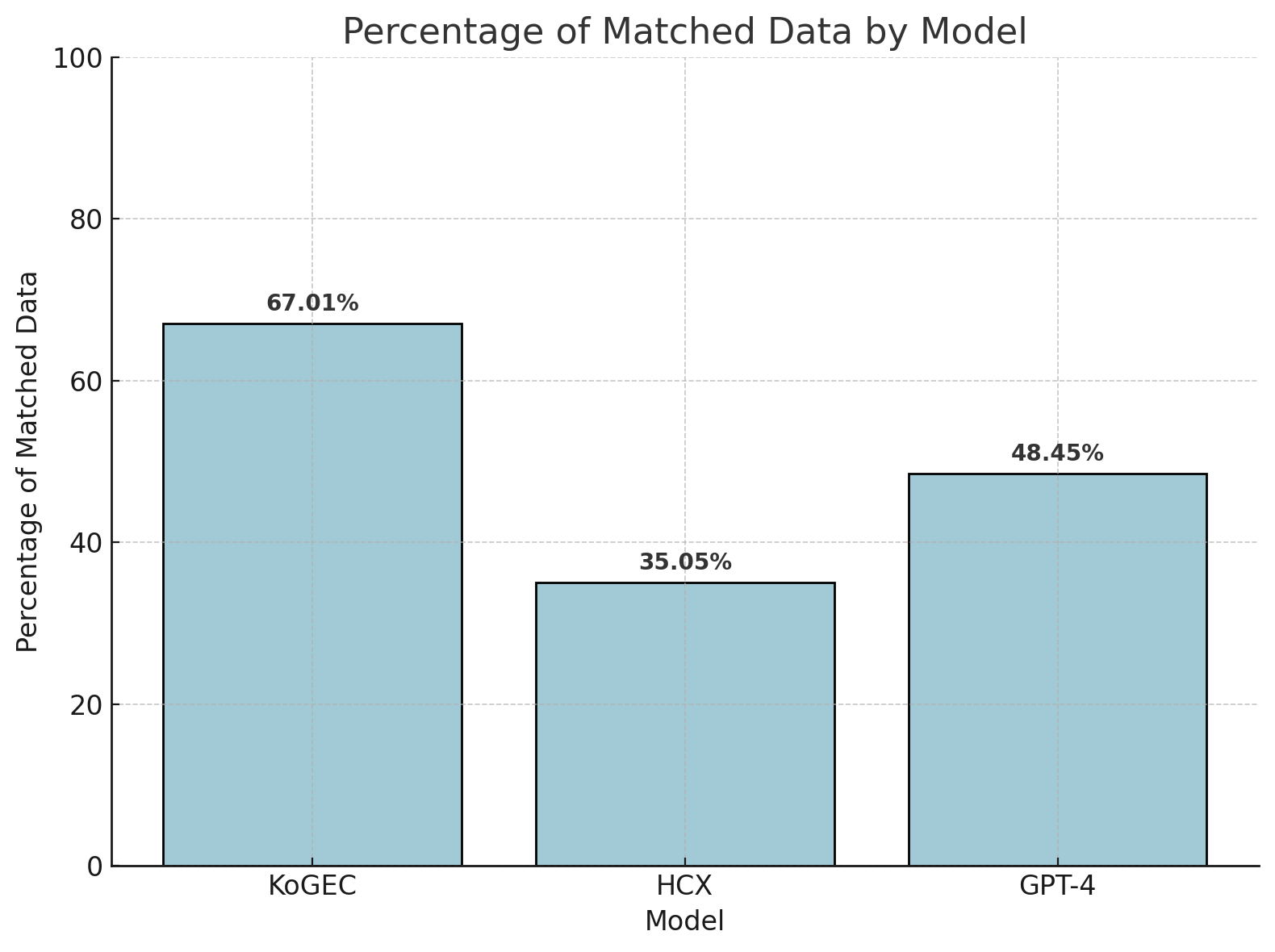}
    \caption{Percentage of matched data by Korean GEC assistant. HCX and GPT-4o have match rates of 35.05\% and 48.45\%, respectively, while KoGEC has a 67.01\% match rate. A breakdown of the error rate by error type is shown in Figure 2.}
    \label{fig:topic_clustering}
\end{figure*}

\begin{figure*}
    \centering
    \includegraphics[width=450pt]{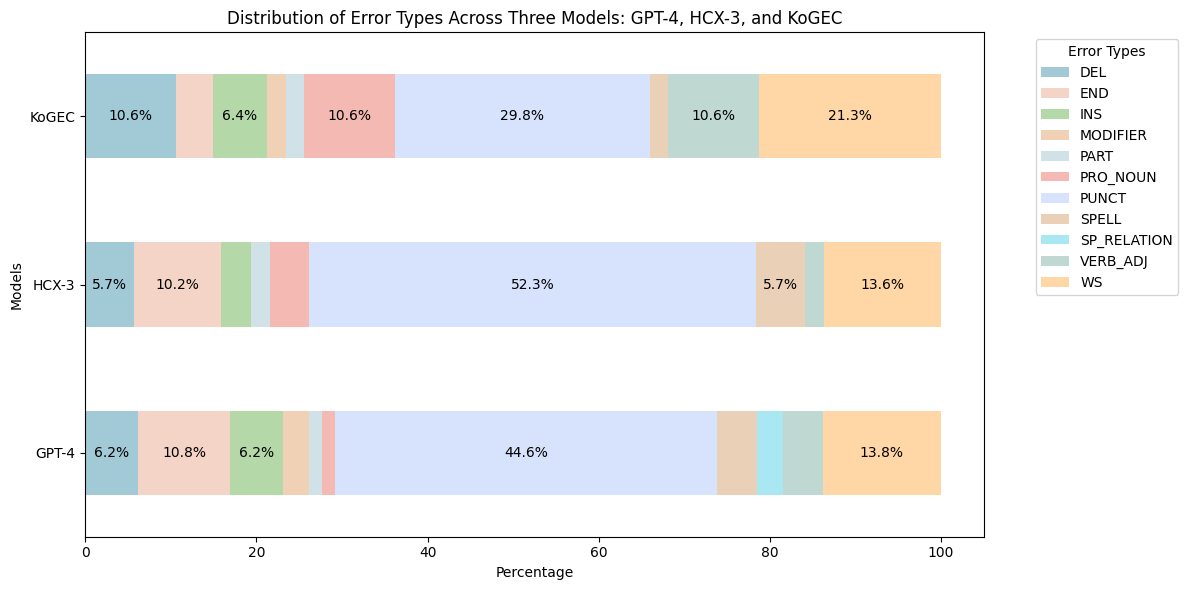}
    \caption{Distribution of error types across three models: GPT-4o, HCX-3, and Ko\--GEC.  Comparative Analysis of Error Type Distribution Across Three Korean Grammatical Error Correction Models.}
    \label{fig:error_distribution}
\end{figure*}

\begin{table}[htbp]
\centering
\begin{tabular}{|l|c|c|c|}
\hline
\textbf{Error Type} & \textbf{GPT-4o} & \textbf{HCX} & \textbf{KoGEC} \\
\hline
DEL        & 6.3\ \ & 5.7\  & 10.6\ \\
END        & 10.9\ & 10.2\ &  4.3\ \\
INS        & 6.3\  &  3.4\ & 6.4\  \\
MODIFIER   & 3.1\  & 0.0\ & 2.1\  \\
PART       & 1.6\  & 2.3\  & 2.1\ \\
PRO\_NOUN  & 1.6\  & 4.5\  & 10.6\ \\
PUNCT      & 43.8\ & 52.3\ & 29.8\ \\
SPELL      & 4.7\  & 5.7\  & 2.1\  \\
SP\_RELATION & 3.1\ &  0.0\  & 0.0\ \\
VERB\_ADJ  & 4.7\  & 2.3\  & 10.6\ \\
WS         & 14.1\ & 13.6\ & 21.3\ \\
\hline
\end{tabular}
\caption{Comparison of Error Types (Unit: \%)}
\label{tab:error-correction-comparison}
\end{table}

GPT-4o and HCX-3 display similar trends, with punctuation (PUNCT) errors dominating at 43.8\% and 52.3\% respectively, followed by word spacing (WS) and ending (END) errors. This suggests these models may be overcompensating punctuation correction at the expense of correcting other error types. Ko\-GEC, in contrast, demonstrates a more balanced error correction profile. While punctuation errors remain the most frequent at 29.8\%, this is significantly lower than the other models. Ko-GEC shows strength in addressing a wider range of error types more evenly: word spacing (WS) errors at 21.3\%, indicating robust performance in a crucial aspect of Korean writing. Equal distribution (10.6\% each) across deletion (DEL), pronoun (PRO\_NOUN) and verb/adjective (VERB\_ADJ) errors, suggesting comprehensive coverage of various grammatical aspects. This balance implies a more comprehensive error correction strategy, potentially offering users a more thorough and nuanced grammatical improvement experience. The model's consistency and versatility to a diverse range of error types with relatively equal emphasis implies a more practical usability for grammatical error correction for native speakers.

\section{Conclusion}
This research introduced KoGEC, a Korean Grammatical Error Correction system that leverages fine-tuned NLLB (No Language Left Behind) models. Our study compared KoGEC's performance against large language models like GPT-4 and HCX-3 using two social media conversation datasets. 
Among the two comparatively small models we tested, we found that the smaller model (NLLB-200-ko-gec-600M) struggled to perform adequately in the Korean GEC task. In contrast, the larger fine-tuned model (NLLB-200-ko-gec-3.3B) not only performed well but outperformed both GPT-4o and HCX-3. The results of this study indicate that model size should be at least 3.3B to achieve good performance, even on specialised NLP tasks such as grammatical error correction.
The evaluation, conducted using BLEU scores and an "LLM as judge" method, demonstrated that KoGEC (specifically the 3.3B model) exhibited a more balanced error correction profile across various error types compared to larger, general-purpose models. This suggests that while raw size is important, targeted fine-tuning on specific tasks can lead to improved performance even with smaller models compared to much larger general-purpose LLMs.
As a practical application of this research, we developed a Chrome extension to make the KoGEC system accessible to users. We aim to create an accessible writing assistant that focuses solely on grammar errors while maintaining the original writing style and purpose. This system is designed to be utilized in low-resource settings for all users.

\section{Further Discussions}
\subsection{Limitations}
In our efforts to investigate ways to further improve our model, we resorted to token vocabulary expansion. We assumed that due to the wide range of languages it covers, the NLLB tokenizer has a relatively shallow coverage of each language. Especially because the Korean language allows, theoretically, for 11,172 baseline syllable letters, the token vocabulary was insufficient to represent all tokens in our dataset. 
Primarily, the ratio of the number of tokens to words per original form and corrected form data were deduced to estimate how well the dataset fit to the NLLB tokenizer. Grammatically accurate data were tokenized to about 1.63 tokens per word, whereas the inaccurate data had a ratio of 2.24 tokens per word. 
To further examine the issue and the tokenizer, we checked for the number of unknown tokens within the entire dataset which added up to 25,831 rows. Having extracted Korean tokens from the NLLB tokenizer, we were able to conclude that NLLB tokenizer vocabulary had 6,789 Korean tokens.
To expand the token vocabulary of the NLLB tokenizer, we trained a separate Sentence Piece tokenizer model on a Korean wikipedia corpus from HuggingFace, where syllable letters that appear more than 5 times within the corpus were assigned as required characters. The trained tokenizer of size 32K was then compared with the original NLLB tokenizer of size 256K to transfer missing tokens and its weights. The tokenizer with expanded vocabulary resulted in 278k tokens in total, which we updated the model to accordingly and trained on the fine-tuning dataset. While we expected a higher performance after ensuring that there were no unknown tokens in the entire corpus, we found that the 3.3B model experienced overfitting by around 14000 steps with batch size of 16, and its performance measured via BLEU score fell behind that of  NLLB-200-ko-gec-3.3B. This must be investigated further, but we suspect that the added tokens were not pre-trained enough. 

\subsection{Future Directions}
Building upon our Korean Grammatical Error Correction system, future research directions present opportunities for expansion and improvement. A primary focus will be on extending our approach to other East Asian languages, particularly Japanese and Chinese. These languages share some structural similarities with Korean, such as complex writing systems and agglutinative or isolating features, which we predict will influence the overall GEC performance. This expansion will not only broaden the applicability of our work but also provide valuable insights into the commonalities and differences in error correction across these linguistically related yet distinct languages.

In parallel with language expansion, we plan to explore the integration of emerging state-of-the-art language models into our GEC framework. Of particular interest is Google's recently released Gemma model, which has shown promising results across various Korean natural language processing tasks. By comparing Gemma's performance against our current NLLB-based approach, we aim to address NLLB's limited token vocabulary.

\bibliography{custom}
\medskip

\appendix

\begin{center}
    \textbf{Appendix}
\end{center}

\section{LLM as judge guide line}
\label{sec:appendix A}
INS: Insertion, where an inserted word adds redundant meaning.\\
Incorrect: 조사조사를 더 많이 해야겠네요. \\
Correct:  조사를 더 많이 해야겠네요.\\
   (We need to do more research.)\\
\medskip
\\
DEL: Deletion, where a deleted word makes the sentence awkward but still understandable. \\
Incorrect: 근데 그때 누 쓰려 하지 않겠냐? \\
Correct:  근데 그때 누구나 쓰려 하지 않겠냐?\\
    (But then who wouldn't want to use it?)\\
\medskip
\\
WS: Word Spacing, violating Korean spacing rules.\\
Incorrect: 오징어 볶음 시키자. \\
Correct:  오징어볶음 시키자.\\
    (Let's order stirfried squid.)\\
\medskip
\\
SPELL: Spelling errors, mainly typing mistakes unrelated to grammar or sentence structure.\\  
Incorrect: 감ㄱ자가 맛있어요.\\
Correct: 감자가 맛있어요. \\
(The potato is delicious.)
\medskip
\\
PUNCT: Punctuation errors, incorrect use of periods, commas, etc.\\
Incorrect: 진짜 한번 가 봐 되게 예뻐.. \\
Correct: 진짜 한번 가 봐. 되게 예뻐.\\
(You should really go see it. It's so pretty.)
\medskip
\\
VERB\_ADJ: Predicate errors, incorrect use of consonants and vowels in standard Korean verbs adjectives.\\
Incorrect: 해시 감자에 기름이 엄청 만아. 어떻해? \\
Correct: 해시 감자에 기름이 엄청 많아. 어떡해?\\
    (The hash browns are so oily. What should I do?)\\
\medskip
\\
PRO\_NOUN (Nominal errors, using non-standard words for nouns, pronouns, numerals, etc.)\\
 Incorrect: 애기랑 나랑 이름이 같다. \\
Correct: 아기랑 나랑 이름이 같다.\\
    (The baby and I have the same name.) \\
\medskip
\\
PART: Particle errors, violating rules for particles that should be combined with preceding nouns.\\
Incorrect: 삼촌가 하와이를 갔다. \\
Correct:  삼촌이 하와이를 갔다.\\
    (My uncle went to hawaii.)\\
\medskip
\\
MODIFIER: Modifier errors.\\
 Incorrect: 외냐하면 예쁘기 때문이다. \\
Correct: 왜냐하면 예쁘기 때문이다. \\
    (Because it's pretty.)\\
\medskip
\\
SP\_RELATION: Sentence coherence errors, changing the structure or meaning of the sentence.\\
Incorrect: 너는결코 혼자야. \\
Correct: 너는 결코 혼자가 아니야. \\
   (You are never alone.)\\
\medskip
\\
END: Ending errors, occurring in tense, connective endings, or final endings.\\
Incorrect: 먹던가 말던가 마음대로 해. \\
Correct : 먹든가 말든가 마음대로 해.\\
   (Whether you eat or not, do as you please.)\\
\medskip
\\
SHORT: Affix errors, occurring in prefixes or suffixes. \\
Incorrect: 솔직이 말해서 출산률이 너무 낮다.\\
Correct : 솔직히 말해서 출산율이 너무 낮다.    \\ 
(To be honest, the birth rate is too low.)\\

\medskip

\section{Korean Orthography Rules}
\label{sec:appendix B}
\begin{itemize}
\item Korean orthography principles are based on writing standard pronunciation while adhering to grammatical rules.
\item In principle, each word in a sentence should be written separately.
\item Loanwords should be written according to the 'Loanword Orthography' rules.
\item When the dependent '-이(\--)' or '\--히\---' follows 'ㄷ', 'ㅌ' endings, even if 'ㄷ', 'ㅌ' sounds like 'ㅈ', 'ㅊ', it should be written as 'ㄷ', 'ㅌ'.  \\ Example: '맏이', not '마지'
\item Among the endings that sound like 'ㄷ', those without a basis for writing as 'ㄷ' should be written as 'ㅅ'. \\ Example: '덧저고리'.
\item The 'ㅖ' in '계', '례', '몌', '폐', '혜' should be written as 'ㅖ' even if it sounds like 'ㅔ'. \\ Example: '계수', not '게수' However, words like '게송' are written according to their original pronunciation.
\item 'ㅢ' in '의' or in syllables starting with a consonant should be written as 'ㅢ' even if it sounds like 'ㅣ'. \\ Example: '의의', not '의이'.
\item When Sino-Korean sounds '녀', '뇨', '뉴', '니' appear at the beginning of a word, they should be written as '여', '요', '유', '이' according to the initial sound law. \\ Example: '여자' [woman], not '녀자'.
\item When Sino-Korean sounds '랴, 려, 례, 료, 류, 리'' appear at the beginning of a word, they should be written as '야, 여, 예, 요, 유, 이' according to the initial sound law. \\ Example: '양심, not '량심'.
\item Nouns should be written separately from particles. \\ Example: 떡이, 떡을, 떡에, 떡도, 떡만
\item The stem and ending of verbs should be written separately. \\ Example: 먹다, 먹고, 먹어, 먹으니.
\item When the last syllable vowel of the stem is 'ㅏ, ㅗ', the ending should be written as '-아', and for other vowels, it should be written as '-어'. \\ Example: 나아, 나아도, 나아서.
\item The particle '요' added after an ending should be written as '요'. \\ Example: 읽어, 읽어요.
\item When '-이' or '-음/-ㅁ' is attached to the stem to form a noun, or '-이' or '-히' is attached to form an adverb, the original form of the stem should be preserved in writing.
\begin{enumerate}
    \item When '-이' is attached to form a noun \\ Example: 길이
\end{enumerate}
\item Words formed by attaching '이' after a noun should be written preserving the original form of the noun.
\begin{enumerate}
    \item When forming an adverb \\ Example: 곳곳이
\end{enumerate}
\item Words formed by attaching a suffix starting with a consonant after a noun or verb stem should be written preserving the original form of the noun or stem. \\ Example: 값지다.
\item Words formed by attaching suffixes '-기-, -리-, -이-, -히-, -구-, -우-, -추-, -으키-, -이키-, -애' to verb stems should be written preserving the original form of the stem. \\ Example: 맡기다
\item When '이' is attached to a root that can take '-하다' or '-거리다' to form a noun, it should be written preserving the original form. \\ Example: 깔쭉이, not 깔쭈기.
\item Verbs formed by attaching '-이다' to onomatopoeic or mimetic roots that can take '-거리다' should be written preserving the original form of the root. \\ Example: 깜짝이다 not 깜짜기다.
\item When '-이' or '-히' is attached to a root that can take '-하다' to form an adverb, or when '-이' is attached to an adverb to intensify its meaning, it should be written preserving the original form of the root or adverb. \\ Example: 급히.
\item Verbs formed by attaching '-하다' or '-없다' should be written preserving '-하다' or '없다'. \\ Example: 딱하다.
\item Words formed by combining two or more words or by attaching a prefix should be written preserving the original form of each component. \\ Example: 국말이
\item Words with clear etymology but unique sound changes should be written as they are pronounced. \\ Example: 할아버지
\item When a word ending with 'ㄹ' is combined with another word and the 'ㄹ' sound is not pronounced, it should be written as it is pronounced. \\ Example: 다달이(달-달-이)
\item When a word ending with 'ㄹ' is combined with another word and the 'ㄹ' sound is pronounced as 'ㄷ', it should be written as 'ㄷ'. \\ Example: 반짇고리(바느질~)
\item 시이소리(linking sound) should be written in the following cases:
\begin{enumerate}
    \item In compound words made of pure Korean words where the first word ends with a vowel \\ Example: 고랫재
\end{enumerate}
\item When two words are combined and a 'ㅂ' or 'ㅎ' sound is added, it should be written as it is pronounced.
\begin{enumerate}
    \item When a 'ㅂ' sound is added \\ Example: 댑싸리
\end{enumerate}
\item When the final vowel of a word is reduced and only the consonant remains, it should be written as a final consonant of the preceding syllable. \\ Example: 기럭아 (기러기야)
\item When a noun and a particle are combined and shortened, they should be written as shortened. \\ Example: 그건(그것은)
\item When '-아/-어, -았-/-었' is combined with stems ending with vowels 'ㅏ, ㅓ', it should be written as shortened. \\ Example: 가(가아)
\item When '-어' follows '이' and is shortened to 'ㅕ', it should be written as shortened. \\ Example: 가져 (가지어)
\item When '-i-' follows stems ending with 'ㅏ, ㅕ, ㅗ, ㅜ, ㅡ' and is shortened to 'ㅐ, ㅖ, ㅚ, ㅟ, ㅢ' respectively, it should be written as shortened. \\ Example: 쌔다 (싸이다)
\item When '-이어' is combined after 'ㅏ, ㅗ, ㅜ, ㅡ' and is shortened, it should be written as shortened. \\ Example: 쌔어 (싸이어)
\item When '-지' is combined with '않-' and becomes '-잖-', or when '-하지' is combined with '않-' and becomes '잖', it should be written as shortened. \\ Example: 그렇잖은 (그렇지 않은)
\item When the 'ㅏ' in the final syllable 'ha' of a stem is reduced and 'ㅎ' combines with the initial sound of the next syllable to form an aspirated sound, it should be written as the aspirated sound. \\ Example: 간편케 (간편하게)
\end{itemize}

Korean Word Spacing Rules:
\begin{itemize}
\item Particles should be attached to the preceding word. \\ Example: 꽃이
\item Dependent nouns should be written separately. \\ Example: 아는 것이 힘이다.
\item Nouns indicating units should be written separately. \\ Example: 한 개
\item When writing numbers, they should be separated in units of 10,000 (man). \\ Example: 십이억 삼천사백오십육만 칠천팔백구십팔.
\item The following words used to connect or list two words should be written separately. \\ Example: 국장 겸 과장
\item When single-syllable words appear consecutively, they can be written together. \\ Example: 좀더
\item Auxiliary verbs should be written separately in principle, but writing them together is also allowed in some cases. \\ Example: 불이 꺼져 간다.(principle)
\item Family names and given names, family names and pen names, etc., should be written together, and titles, official positions, etc., added to these should be written separately. \\ Example: 김양수.
\item Proper nouns other than personal names should be written separately by word in principle but can be written separately by unit. \\ Example: 대한 중학교.
\item Technical terms should be written separately by word in principle but can be written together. \\ Example: 만성 골수성 백혈병 (principle)
\item For adverbs, if the final syllable clearly sounds only as 'i', it should be written as '-이', and if it sounds only as 'hi' or as either 'i' or 'hi', it should be written as '-hi'. \\ Example: 가붓이
\item In Sino-Korean words, those that are pronounced in both their original sound and colloquial sound should be written according to each pronunciation. \\ Example: 승낙(pronounced in original sound)
\item The following endings should be written with unaspirated sounds. \\ Example: -(으)ㄹ거나
\item The following suffixes should be written with tense sounds. \\ Example: 심부름꾼.
\item The following words that were previously written in two different ways should now be written in one way. \\ Example: 맞추다(입을 맞춘다, 양복을 맞춘다).
\item Endings indicating past events should be written as '-든지', '-던' instead of '-던지, -던'. \\ Example: 춥더라.
\item The following words should be written separately. \\ Example: 가름, 갈음
\end{itemize}

\end{document}